\documentclass[10pt,conference]{IEEEtran}

\usepackage[T1]{fontenc}
\usepackage[utf8]{inputenc}

\usepackage{amsmath}
\usepackage{amssymb}
\usepackage{amsfonts}
\usepackage{mathtools}
\usepackage{bm}

\usepackage{graphicx}
\usepackage{booktabs}
\usepackage{multirow}
\usepackage{array}
\usepackage{makecell}
\usepackage{tabularx}
\usepackage{stfloats}
\usepackage{algorithm}
\usepackage{algorithmic}

\usepackage{acro}

\acsetup{
  first-style = short-long,
  list/display = used,
  make-links = true
}

\DeclareAcronym{a2q}{
  short = A2Q,
  long  = accumulator-aware quantisation
}

\DeclareAcronym{smt}{
  short = SMT,
  long  = satisfiability modulo theories
}

\DeclareAcronym{ai}{
  short = AI,
  long  = artificial intelligence
}

\DeclareAcronym{ann}{
  short = ANN,
  long  = artificial neural network
}

\DeclareAcronym{dnn}{
  short = DNN,
  long  = deep neural network
}

\DeclareAcronym{cnn}{
  short = CNN,
  long  = convolutional neural network
}

\DeclareAcronym{mlp}{
  short = MLP,
  long  = multi-layer perceptron
}

\DeclareAcronym{vit}{
  short = ViT,
  long  = vision transformer
}

\DeclareAcronym{fp32}{
  short = FP32,
  long  = 32-bit floating point
}

\DeclareAcronym{fpga}{
  short = FPGA,
  long  = field-programmable gate array
}

\DeclareAcronym{asic}{
  short = ASIC,
  long  = application-specific integrated circuit
}

\DeclareAcronym{mac}{
  short = MAC,
  long  = multiply--accumulate
}

\DeclareAcronym{alu}{
  short = ALU,
  long  = arithmetic logic unit
}

\DeclareAcronym{lut}{
  short = LUT,
  long  = look-up table
}

\DeclareAcronym{ptq}{
  short = PTQ,
  long  = post-training quantisation
}

\DeclareAcronym{qat}{
  short = QAT,
  long  = quantisation-aware training
}

\DeclareAcronym{ste}{
  short = STE,
  long  = straight-through estimator
}

\DeclareAcronym{qnn}{
  short = QNN,
  long  = quantised neural network
}

\DeclareAcronym{lqnn}{
  short = LQNN,
  long  = Lyapunov-stabilised quantised neural network
}

\DeclareAcronym{fxp}{
  short = FXP,
  long  = fixed-point
}

\DeclareAcronym{int8}{
  short = INT8,
  long  = 8-bit integer
}

\DeclareAcronym{int16}{
  short = INT16,
  long  = 16-bit integer
}

\DeclareAcronym{qformat}{
  short = Q-format,
  long  = fixed-point integer--fractional format
}

\DeclareAcronym{tc}{
  short = TC,
  long  = two's-complement
}

\DeclareAcronym{wrap}{
  short = WRAP,
  long  = overflow wrapping
}

\DeclareAcronym{sat}{
  short = SAT,
  long  = saturation
}

\DeclareAcronym{ce}{
  short = CE,
  long  = cross-entropy
}

\DeclareAcronym{mse}{
  short = MSE,
  long  = mean squared error
}

\DeclareAcronym{mc}{
  short = MC,
  long  = Monte Carlo
}

\DeclareAcronym{std}{
  short = STD,
  long  = standard deviation
}

\DeclareAcronym{mnist}{
  short = MNIST,
  long  = Modified National Institute of Standards and Technology
}

\DeclareAcronym{relu}{
  short = ReLU,
  long  = rectified linear unit
}

\DeclareAcronym{pdf}{
  short = PDF,
  long  = portable document format
}

\DeclareAcronym{csv}{
  short = CSV,
  long  = comma-separated values
}

\usepackage[hidelinks]{hyperref}

\graphicspath{{figures/}}

\usepackage[
backend=biber,
style=numeric,
sorting=ynt
]{biblatex}

\usepackage{amsthm}

\begin{document}
\title{Lyapunov-Guided Training for Hardware-Safe Neural Networks Under Fixed-Point Arithmetic}

\author{\IEEEauthorblockN{Anis Hamadouche}
\IEEEauthorblockA{\textit{The School of Engineering \& Physical Sciences} \\
\textit{Heriot-Watt University}\\
Edinburgh EH14 4AS, UK \\
anis.hamadouche@hw.ac.uk}

\and

\IEEEauthorblockN{Amir Hussain}
\IEEEauthorblockA{\textit{SDAIA-KFUPM Joint Research Centre for Artificial Intelligence} \\
\textit{King Fahd University of Petroleum and Minerals}\\
Dhahran, Saudi Arabia  \\
amir.hussain@kfupm.edu.sa}
}

\maketitle
\thispagestyle{plain}
\pagestyle{plain}

\begin{abstract}
Low-precision neural networks are attractive for resource-constrained hardware, but fixed-point arithmetic introduces failure modes that are often hidden by idealised quantisation models. In particular, two's-complement overflow wrapping can corrupt hidden activations by changing both their magnitude and sign, leading to unstable numerical error propagation and severe accuracy degradation. This paper proposes a Lyapunov-stabilised quantisation framework for low-precision neural networks operating under hardware-style wrapping arithmetic. The hidden-state energy is monitored through a layerwise Lyapunov function, and a monotone projection is applied to enforce bounded and non-increasing state evolution across depth. The method is evaluated on MNIST using a compact patch-based transformer under post-training quantisation and quantisation-aware training with fixed-point bit-widths from \(4\) to \(16\) bits. Monte Carlo results show that unconstrained wrapped quantisation-aware training collapses to near-chance accuracy across \(6\)--\(16\) bits, with activation overflow rates exceeding \(11\%\). In contrast, the proposed monotone Lyapunov projection suppresses activation overflow to below \(0.012\%\) and restores stable low-precision learning, achieving \(86.55\%\pm0.65\%\) accuracy at \(12\) bits. These results demonstrate that Lyapunov-based state control can act as a hardware-aware stabilisation mechanism for reliable fixed-point neural inference and training.
\end{abstract}


\IEEEpeerreviewmaketitle
\newcommand\scalemath[2]{\scalebox{#1}{\mbox{\ensuremath{\displaystyle #2}}}}
\section{Introduction}
\label{sec:introduction}

The deployment of deep neural networks on edge devices, embedded processors, field-programmable gate arrays, and custom accelerators has made low-precision arithmetic a central topic in hardware-aware machine learning. Quantisation reduces memory footprint, bandwidth, arithmetic cost, and energy consumption by representing weights, activations, or accumulators using fewer bits than conventional floating-point arithmetic. These benefits are particularly important for real-time and resource-constrained applications, where full-precision inference may be too expensive in terms of latency, silicon area, or power consumption. Consequently, post-training quantisation (PTQ) and quantisation-aware training (QAT) have become standard techniques for mapping neural networks onto efficient low-precision hardware \cite{gholami2021survey,weng2021quantization,wei2024advances}.

Most quantisation studies focus on rounding error, clipping error, calibration, mixed precision, or the accuracy degradation caused by reducing the number of representable values. However, practical fixed-point hardware introduces an additional failure mode: arithmetic overflow. In saturating arithmetic, an overflowed value is clipped to the largest or smallest representable value. In wrapping arithmetic, which follows two's-complement behaviour, an overflowed value wraps around the finite integer range. This means that a large positive value may become negative, and a large negative value may become positive. Such behaviour is efficient in hardware but can be numerically destructive. In quantised neural networks, overflow in activations or accumulators can corrupt hidden-state trajectories and produce abrupt changes in the represented computation \cite{baranowski2020fixedpoint,colbert2023a2q}.

This issue is especially important for deep residual and transformer-like architectures. These models can be interpreted as depth-indexed dynamical systems, where each layer updates a hidden state through a nonlinear transformation. In such a view, unstable hidden-state growth across depth is not merely a numerical inconvenience; it is a dynamical instability. If the hidden-state energy grows beyond the representable fixed-point range, overflow wrapping can produce discontinuous sign and magnitude errors. These errors may then propagate through subsequent layers, producing catastrophic degradation even when the nominal bit-width appears sufficiently large. Therefore, robust low-precision learning requires not only quantisation-aware optimisation, but also explicit control of the internal state trajectory.

Lyapunov methods provide a natural framework for analysing and controlling such layerwise dynamics. A Lyapunov function assigns a non-negative scalar energy to the system state and can be used to certify boundedness or monotonic decrease along the system trajectory. Recent work has explored connections between residual networks, stability, and dynamical systems, suggesting that depth-wise stability can influence optimisation, generalisation, and robustness \cite{nar2018residual}. This motivates the use of Lyapunov-inspired constraints for low-precision neural networks: if the hidden-state energy can be kept inside a safe set, the network is less likely to enter overflow-prone regions of the fixed-point state space.

In this paper, we propose a Lyapunov-stabilised quantisation framework for neural networks operating under hardware-style fixed-point wrapping arithmetic. The key idea is to monitor the normalised hidden-state energy at each layer and apply a monotone projection that enforces
\begin{equation}
V(h^{\ell+1})
\leq
\min\{V(h^\ell),V_{\max}\},
\end{equation}
where \(V(h^\ell)\) denotes the layerwise Lyapunov energy and \(V_{\max}\) defines a safe energy threshold. Unlike a soft regularisation term alone, the projection directly constrains the hidden state and prevents the network from entering regions where fixed-point wraparound is likely to occur. In quantised models, the projection is applied after fixed-point write-back, so that quantisation-induced overflow is explicitly accounted for.

The proposed method is evaluated on MNIST using a compact patch-based transformer. We compare full-precision training, PTQ, QAT, and their Lyapunov-projected counterparts across fixed-point bit-widths from \(4\) to \(16\) bits. The experiments use signed fixed-point arithmetic with two's-complement wrapping, rather than idealised clipping. Monte Carlo results show that unconstrained wrapped QAT collapses to near-random accuracy across \(6\)--\(16\) bits, with activation overflow rates exceeding \(11\%\). In contrast, QAT with monotone Lyapunov projection suppresses overflow to approximately zero and recovers stable low-precision learning, achieving \(86.55\%\pm0.65\%\) accuracy at \(12\) bits.

The main contributions of this work are as follows. First, we formulate fixed-point neural-network quantisation with overflow wrapping as a hidden-state stability problem. Second, we introduce a monotone Lyapunov projection mechanism that enforces bounded and non-increasing layerwise energy. Third, we demonstrate that the proposed projection significantly improves PTQ and QAT robustness under hardware-style wrapping arithmetic. Finally, we provide empirical evidence that activation overflow, rather than bit-width alone, can be a dominant failure mode in low-precision neural networks.






\section{Related Work}
\label{sec:related_work}

\subsection{Neural Network Quantisation}

Quantisation is one of the main techniques for deploying neural networks on resource-constrained hardware. By reducing the precision of weights, activations, and sometimes intermediate accumulators, quantisation can reduce memory footprint, bandwidth, arithmetic cost, and energy consumption. Existing methods are commonly divided into \ac{ptq} and \ac{qat}. \ac{ptq} applies quantisation to a trained full-precision model, usually with calibration data but without full retraining. \ac{qat}, by contrast, simulates quantisation during training or fine-tuning, often using a straight-through estimator to approximate gradients through non-differentiable rounding operations. Surveys such as \cite{gholami2021survey,nagel2021whitepaper,weng2021quantization,rokh2022comprehensive} provide comprehensive overviews of uniform and non-uniform quantisation, calibration, clipping, mixed precision, and hardware-aware deployment.

A particularly influential direction is integer-only inference. Jacob et al. \cite{jacob2018quantization} proposed a quantisation scheme that enables inference using integer arithmetic, showing that low-precision execution can preserve accuracy while improving hardware efficiency. This line of work has strongly influenced deployment pipelines for mobile and edge devices, where integer arithmetic is typically more efficient than floating-point computation. Subsequent work extended these ideas to more complex networks, including transformer-based models, where integer-only or mixed-precision execution is used to reduce inference cost.

Most quantisation methods focus primarily on quantisation noise, rounding error, scale selection, clipping thresholds, and the mismatch between full-precision and low-precision representations. In these settings, saturation or clipping is often assumed when values exceed the representable range. However, practical fixed-width arithmetic may instead use two's-complement wrapping, where overflowed values re-enter the representable range with altered sign and magnitude. This distinction is important because wrapping is not merely a bounded approximation error; it can introduce discontinuous state corruption. Therefore, fixed-point neural inference under wrapping arithmetic requires stability analysis beyond standard quantisation-error models.

\subsection{Overflow-Aware and Accumulator-Aware Quantisation}

Overflow has received increasing attention in low-precision neural-network deployment. In integer-based accelerators, the output of multiply--accumulate operations may require more bits than the input operands. If the accumulator width is insufficient, overflow can corrupt the computation even when weights and activations are individually representable. De Bruin et al. \cite{debruin2020accumulator} studied quantisation for integer platforms without wide accumulators, highlighting the practical importance of accumulator constraints.

More recently, Colbert et al. proposed \ac{a2q}, an accumulator-aware quantisation method with guaranteed overflow avoidance \cite{colbert2023a2q}. Their approach constrains the \(\ell_1\)-norm of model weights according to derived accumulator bit-width bounds, thereby preventing accumulator overflow during inference. This work is closely related in motivation to the present paper, because both approaches treat overflow as a first-order hardware failure mode rather than a minor numerical artefact.

However, our focus differs in two ways. First, \ac{a2q} primarily targets accumulator overflow through weight constraints, whereas the present work studies activation-state overflow and hidden-state dynamics across network depth. Second, we explicitly consider two's-complement wrapping behaviour, where overflowed values can change sign and corrupt the state trajectory. The proposed Lyapunov projection therefore acts directly on the hidden state, enforcing a safe energy set after quantised write-back.

\subsection{Fixed-Point Arithmetic and Sound Quantised Implementations}

Fixed-point arithmetic is attractive for hardware because it offers predictable area, latency, and power characteristics. Nevertheless, choosing a fixed-point format requires balancing integer range, fractional precision, and overflow risk. Formal and verification-oriented work has studied fixed-point arithmetic from the perspective of soundness and implementation correctness. Baranowski et al. \cite{baranowski2020smt} developed an \ac{smt} theory of fixed-point arithmetic, supporting formal reasoning about finite-word-length computations. Other work has considered mixed fixed-point quantisation with explicit overflow-freedom constraints, often using optimisation or verification methods to choose precision assignments automatically.

These approaches are complementary to the proposed method. Formal methods can prove that a given implementation satisfies certain numerical constraints, while mixed-precision search can allocate bit-widths across layers. In contrast, our method modifies the network dynamics by projecting the hidden state into a Lyapunov-safe set. The goal is not only to choose a sufficient precision, but also to prevent unstable hidden-state evolution from entering overflow-prone regions of the fixed-point state space.

\subsection{Stability Views of Deep Neural Networks}

Deep residual networks and transformer-like architectures can be interpreted as depth-indexed dynamical systems. In this interpretation, each layer applies a nonlinear update to a hidden state, and the network output is obtained after a finite number of state transitions. This viewpoint has motivated the use of tools from dynamical systems, numerical analysis, and control theory to study deep-network stability.

Nar and Sastry \cite{nar2018residual} analysed residual networks from a Lyapunov-stability perspective and connected residual parametrisation to stable gradient-based optimisation and generalisation behaviour. Related work on neural ordinary differential equations and stable neural dynamical systems has also explored how stability constraints can improve robustness and long-horizon behaviour \cite{kang2021sodef,rodriguez2022lyanet}. In control-oriented settings, neural Lyapunov functions have been used to certify stability or safety of learned controllers \cite{richards2018lyapunov,dai2021lyapunov}.

The present paper adopts a similar stability viewpoint, but applies it to low-precision hardware execution. Instead of using Lyapunov functions only to analyse continuous-time dynamics or closed-loop control systems, we use a layerwise Lyapunov energy to constrain the hidden-state trajectory of a quantised neural network. This is particularly useful under fixed-point wrapping arithmetic, where unstable energy growth can lead directly to overflow-induced state corruption.

\subsection{Positioning of This Work}

Existing quantisation methods mainly address how to approximate full-precision weights and activations using fewer bits. Overflow-aware methods, such as accumulator-aware quantisation, explicitly address numerical range limits but primarily focus on accumulator safety. Stability-oriented neural-network methods study boundedness, robustness, or convergence, but usually not under realistic fixed-point wrapping arithmetic.

This work connects these directions by treating low-precision neural inference and training as a dynamical stability problem. The proposed monotone Lyapunov projection enforces
\begin{equation}
V(h^{\ell+1}) \leq \min\{V(h^\ell),V_{\max}\},
\end{equation}
thereby bounding the layerwise hidden-state energy before overflow wrapping can dominate the computation. Unlike clipping-based quantisation, the proposed method does not simply saturate individual scalar values. Instead, it constrains the full hidden-state energy and therefore controls the state trajectory of the network across depth.

The resulting framework is hardware-aware because it explicitly models fixed-point write-back and two's-complement wrapping. It is also stability-aware because it constrains the hidden-state dynamics rather than only the quantisation error. This combination distinguishes the proposed method from conventional \ac{ptq}, standard \ac{qat}, and accumulator-only overflow-avoidance techniques.

\section{System Model and Proposed Method}
\label{sec:system_model_method}

\subsection{Depth-Indexed Neural Dynamics}
\label{subsec:depth_indexed_dynamics}

Consider a neural network with \(L\) hidden layers. Let \(h^\ell\in\mathbb{R}^{T\times D}\) denote the hidden state at layer \(\ell\), where \(T\) is the number of tokens and \(D\) is the embedding dimension. For a residual or transformer-like block, the layerwise update can be written as
\begin{equation}
h^{\ell+1}
=
h^\ell + F_\ell(h^\ell;\theta_\ell),
\qquad
\ell=0,\ldots,L-1,
\label{eq:residual_dynamics}
\end{equation}
where \(F_\ell(\cdot;\theta_\ell)\) denotes the nonlinear transformation implemented by the \(\ell\)-th block and \(\theta_\ell\) contains the associated trainable parameters. The final prediction is obtained from
\begin{equation}
\hat{y}
=
g(h^L;\theta_g),
\label{eq:classifier_head}
\end{equation}
where \(g(\cdot)\) is the classification head.

This representation treats depth as a discrete-time dynamical axis. Under this viewpoint, the hidden states
\[
h^0,h^1,\ldots,h^L
\]
form a state trajectory. Stability of this trajectory is important for low-precision hardware because excessive hidden-state growth can push activations outside the representable fixed-point range. In wrapping arithmetic, such an overflow does not merely clip the value; it can change the sign and magnitude of the represented state.

\subsection{Fixed-Point Quantisation with Overflow Wrapping}
\label{subsec:fixed_point_wrapping}

We consider signed fixed-point quantisation with \(B\) total bits. The format consists of one sign bit, \(I\) integer bits, and \(F\) fractional bits, such that
\begin{equation}
B = 1 + I + F .
\end{equation}
For a real-valued scalar \(x\), the corresponding fixed-point integer is
\begin{equation}
q
=
\operatorname{round}\left(2^F x\right).
\label{eq:fixed_point_integer}
\end{equation}
The signed integer range is
\begin{equation}
q_{\min}=-2^{B-1},
\qquad
q_{\max}=2^{B-1}-1.
\end{equation}

In saturation arithmetic, \(q\) would be clipped to the interval
\([q_{\min},q_{\max}]\). In this work, however, we explicitly model two's-complement wrapping. The wrapped integer is given by
\begin{equation}
q_{\mathrm{wrap}}
=
\left((q-q_{\min}) \bmod 2^B\right)+q_{\min}.
\label{eq:wrap_integer}
\end{equation}
The dequantised fixed-point value is then
\begin{equation}
\mathcal{Q}_B(x)
=
\frac{q_{\mathrm{wrap}}}{2^F}.
\label{eq:dequantised_value}
\end{equation}

Equation~\eqref{eq:wrap_integer} captures hardware-style overflow. For example, in signed \(8\)-bit integer arithmetic, \(127+1\) wraps to \(-128\), while \(-128-1\) wraps to \(127\). Therefore, overflow may introduce a large discontinuity in the represented value. This motivates controlling the hidden-state trajectory before it enters overflow-prone regions.

For a tensor \(h^\ell\), quantisation is applied elementwise:
\begin{equation}
\widetilde{h}^{\ell}
=
\mathcal{Q}_B(h^\ell).
\label{eq:tensor_quantisation}
\end{equation}
During quantisation-aware training, the forward pass uses \(\mathcal{Q}_B(\cdot)\), while the backward pass uses the straight-through estimator,
\begin{equation}
\frac{\partial \mathcal{Q}_B(x)}{\partial x}
\approx
1.
\label{eq:ste}
\end{equation}

\subsection{Layerwise Lyapunov Energy}
\label{subsec:layerwise_lyapunov_energy}

To monitor the internal state trajectory, we define the normalised layerwise Lyapunov energy
\begin{equation}
V(h^\ell)
=
\frac{1}{TD}
\|h^\ell\|_F^2.
\label{eq:lyapunov_energy}
\end{equation}
This energy measures the average hidden-state magnitude per token and feature dimension. A stable low-precision trajectory should remain inside a safe set
\begin{equation}
\mathcal{S}
=
\left\{
h : V(h)\leq V_{\max}
\right\},
\label{eq:safe_set}
\end{equation}
where \(V_{\max}>0\) is a prescribed energy threshold.

The threshold is estimated from the initial token energy distribution on the training set. Let \(V(h^0_i)\) denote the initial energy of the \(i\)-th training sample. We set
\begin{equation}
V_{\max}
=
\gamma
\,
\operatorname{Quantile}_{p}
\left(
\{V(h^0_i)\}_{i=1}^{N}
\right),
\label{eq:threshold_definition}
\end{equation}
where \(p\in(0,1)\) is a high quantile and \(\gamma>1\) is a safety margin. In the experiments, \(p=0.99\) and \(\gamma=1.20\).

\subsection{Monotone Lyapunov Projection}
\label{subsec:monotone_projection}

The proposed method constrains the hidden-state trajectory using a monotone Lyapunov projection. Given an unconstrained updated state \(z^{\ell+1}\), we project it onto the energy ball
\begin{equation}
V(h^{\ell+1})
\leq
\min\{V(h^\ell),V_{\max}\}.
\label{eq:monotone_condition}
\end{equation}
Define the target energy
\begin{equation}
\bar{V}^{\ell+1}
=
\min\{V(h^\ell),V_{\max}\}.
\label{eq:target_energy}
\end{equation}
The projected state is
\begin{equation}
h^{\ell+1}
=
\Pi_{\bar{V}^{\ell+1}}(z^{\ell+1})
=
\begin{cases}
z^{\ell+1},
&
V(z^{\ell+1}) \leq \bar{V}^{\ell+1},
\\[2mm]
z^{\ell+1}
\sqrt{
\frac{\bar{V}^{\ell+1}}
{V(z^{\ell+1})+\epsilon}
},
&
V(z^{\ell+1}) > \bar{V}^{\ell+1},
\end{cases}
\label{eq:projection_operator}
\end{equation}
where \(\epsilon>0\) is a small numerical constant.

The projection in \eqref{eq:projection_operator} enforces
\begin{equation}
V(h^{\ell+1})
\leq
\bar{V}^{\ell+1}
=
\min\{V(h^\ell),V_{\max}\}.
\label{eq:projection_guarantee}
\end{equation}
Therefore, the projected trajectory satisfies
\begin{equation}
V(h^{\ell+1})
\leq
V(h^\ell),
\qquad
V(h^{\ell+1})
\leq
V_{\max}.
\label{eq:bounded_nonincreasing}
\end{equation}
Thus, the layerwise energy is both bounded and non-increasing across depth.

\subsection{Quantised Lyapunov-Stabilised Forward Pass}
\label{subsec:quantised_forward_pass}

The order of operations is important under wrapping arithmetic. If projection is applied before quantisation, then fixed-point write-back may reintroduce overflow and violate the safe-set constraint. Therefore, the proposed quantised forward pass applies projection after quantised write-back.

For each layer, the unconstrained full-precision update is first computed as
\begin{equation}
z^{\ell+1}
=
h^\ell
+
F_\ell(h^\ell;\theta_\ell).
\label{eq:unconstrained_update}
\end{equation}
The state is then quantised using wrapping arithmetic:
\begin{equation}
\widetilde{z}^{\ell+1}
=
\mathcal{Q}_B(z^{\ell+1}).
\label{eq:wrapped_state}
\end{equation}
The Lyapunov projection is applied to the wrapped state:
\begin{equation}
h^{\ell+1}
=
\Pi_{\bar{V}^{\ell+1}}
\left(
\widetilde{z}^{\ell+1}
\right).
\label{eq:project_after_quantisation}
\end{equation}
If the projected state is written back to fixed-point memory, it is quantised again:
\begin{equation}
\widehat{h}^{\ell+1}
=
\mathcal{Q}_B(h^{\ell+1}).
\label{eq:projected_writeback}
\end{equation}
A repair projection may then be applied:
\begin{equation}
h^{\ell+1}
=
\Pi_{\bar{V}^{\ell+1}}
\left(
\widehat{h}^{\ell+1}
\right),
\label{eq:repair_projection}
\end{equation}
which ensures that fixed-point write-back does not break the Lyapunov constraint.

\subsection{Training Objective}
\label{subsec:training_objective}

Let \(\mathcal{D}=\{(x_i,y_i)\}_{i=1}^{N}\) denote the training set. The standard task loss is the cross-entropy loss
\begin{equation}
\mathcal{L}_{\mathrm{CE}}(\theta)
=
-\frac{1}{N}
\sum_{i=1}^{N}
\log
p_\theta(y_i|x_i).
\label{eq:cross_entropy}
\end{equation}
For full-precision training with Lyapunov regularisation, we may use the penalised objective
\begin{equation}
\mathcal{L}(\theta)
=
\mathcal{L}_{\mathrm{CE}}(\theta)
+
\lambda
\sum_{\ell=1}^{L}
\left[
\max\{0,V(h^\ell)-V_{\max}\}
\right]^2,
\label{eq:lyapunov_penalised_loss}
\end{equation}
where \(\lambda\geq0\) controls the penalty strength.

For the projected models, the hard Lyapunov constraint is enforced directly by \eqref{eq:projection_operator}. In this case, the optimisation objective remains
\begin{equation}
\min_{\theta}
\mathcal{L}_{\mathrm{CE}}(\theta),
\label{eq:projected_training_objective}
\end{equation}
subject to the layerwise projected dynamics
\begin{equation}
h^{\ell+1}
=
\Pi_{\bar{V}^{\ell+1}}
\left(
\mathcal{Q}_B
\left(
h^\ell+F_\ell(h^\ell;\theta_\ell)
\right)
\right).
\label{eq:constrained_layer_dynamics}
\end{equation}

\subsection{Overflow and Stability Metrics}
\label{subsec:metrics}

To quantify the behaviour of each model, we report accuracy, loss, maximum layerwise energy, mean state change, projection rate, and activation overflow rate. The maximum energy is
\begin{equation}
V_{\max}^{\mathrm{obs}}
=
\max_{\ell}
V(h^\ell).
\label{eq:observed_max_energy}
\end{equation}
The mean layerwise state change is
\begin{equation}
\Delta_{\mathrm{mean}}
=
\frac{1}{L}
\sum_{\ell=0}^{L-1}
\frac{1}{N}
\sum_{i=1}^{N}
\sqrt{
\frac{1}{TD}
\|h^{\ell+1}_i-h^\ell_i\|_F^2
}.
\label{eq:mean_state_change}
\end{equation}
The projection rate at layer \(\ell\) is the fraction of samples for which
\[
V(\widetilde{z}^{\ell+1}) > \bar{V}^{\ell+1}.
\]
The activation overflow rate is the fraction of fixed-point activation values for which
\[
q < q_{\min}
\quad
\text{or}
\quad
q > q_{\max}
\]
before wrapping is applied.

\begin{algorithm}[t]
\caption{Quantised Monotone Lyapunov Forward Pass}
\label{alg:quantised_lyapunov_forward}
\begin{algorithmic}[1]
\REQUIRE Input \(h^0\), parameters \(\{\theta_\ell\}_{\ell=0}^{L-1}\), bit-width \(B\), safe threshold \(V_{\max}\)
\FOR{\(\ell=0,\ldots,L-1\)}
    \STATE Compute unconstrained update:
    \[
    z^{\ell+1}=h^\ell+F_\ell(h^\ell;\theta_\ell)
    \]
    \STATE Apply fixed-point wrapping:
    \[
    \widetilde{z}^{\ell+1}=\mathcal{Q}_B(z^{\ell+1})
    \]
    \STATE Set target energy:
    \[
    \bar{V}^{\ell+1}=\min\{V(h^\ell),V_{\max}\}
    \]
    \STATE Project onto the Lyapunov-safe set:
    \[
    h^{\ell+1}
    =
    \Pi_{\bar{V}^{\ell+1}}
    \left(
    \widetilde{z}^{\ell+1}
    \right)
    \]
    \STATE Quantise projected state for write-back:
    \[
    \widehat{h}^{\ell+1}
    =
    \mathcal{Q}_B(h^{\ell+1})
    \]
    \STATE Repair projection:
    \[
    h^{\ell+1}
    =
    \Pi_{\bar{V}^{\ell+1}}
    \left(
    \widehat{h}^{\ell+1}
    \right)
    \]
\ENDFOR
\RETURN \(h^L\)
\end{algorithmic}
\end{algorithm}

\section{Experimental Results}
\label{sec:experimental_results}


We evaluate the proposed Lyapunov-stabilised low-precision transformer on the MNIST classification task. The network is a compact patch-based transformer in which each \(28\times 28\) image is divided into non-overlapping \(7\times 7\) patches, producing \(16\) tokens per image. Each token is embedded into a \(64\)-dimensional representation and processed by a \(4\)-layer transformer-like stack. The classifier is applied to the mean-pooled final token representation.

The experiments compare full-precision training, post-training quantisation, and quantisation-aware training under signed fixed-point arithmetic. The fixed-point format uses \(B\) total bits, consisting of one sign bit, a fixed number of integer bits, and the remaining fractional bits. Unless otherwise stated, the weight format uses one integer bit and the activation format uses two integer bits. The quantised integer is computed as
\begin{equation}
q = \operatorname{round}\left(x2^{F}\right),
\end{equation}
where \(F\) is the number of fractional bits. To emulate hardware overflow, signed two's-complement wrapping is used:
\begin{equation}
q_{\mathrm{wrap}}
=
\left((q-q_{\min}) \bmod 2^B\right)+q_{\min},
\end{equation}
where \(q_{\min}=-2^{B-1}\). The dequantised value is then
\begin{equation}
\hat{x}=\frac{q_{\mathrm{wrap}}}{2^F}.
\end{equation}
This differs from idealised clipping or saturation because overflow may change both the magnitude and sign of the represented value.

For Lyapunov monitoring, we define the normalised hidden-state energy
\begin{equation}
V(h^\ell)
=
\frac{1}{TD}
\|h^\ell\|_F^2,
\end{equation}
where \(T\) is the number of tokens and \(D\) is the embedding dimension. The monotone Lyapunov projection enforces
\begin{equation}
V(h^{\ell+1})
\leq
\min\left\{V(h^\ell),V_{\max}\right\},
\end{equation}
where \(V_{\max}\) is estimated from the \(0.99\)-quantile of the initial training-set token energy with a safety margin of \(1.20\). In the quantised projected models, the state is first quantised using wrapping arithmetic and then projected. An additional repair projection is applied after quantised write-back to prevent fixed-point write-back from violating the Lyapunov constraint.

All reported results are averaged over three Monte Carlo runs with different random seeds. The training set contains \(10{,}000\) images, the validation set \(2{,}000\) images, and the test set \(2{,}000\) images. The full-precision models are trained for \(5\) epochs using AdamW with learning rate \(2\times 10^{-3}\), while quantisation-aware fine-tuning uses learning rate \(5\times 10^{-4}\). The bit-width sweep considers
\[
B\in\{4,6,8,10,12,16\}.
\]

\subsection{Results and Discussion}
\label{subsec:results_discussion}

Table~\ref{tab:mnist_mc_summary} summarises the Monte Carlo test results. The full-precision baseline achieves the highest accuracy, \(89.15\%\pm0.64\%\), but its hidden-state energy grows substantially, reaching a maximum normalised energy of \(6.2634\). By contrast, the full-precision model with monotone Lyapunov projection achieves \(86.95\%\pm0.85\%\), while reducing the maximum energy to \(0.3595\). Hence, the Lyapunov constraint introduces a modest accuracy cost of approximately \(2.2\) percentage points, but reduces the maximum hidden-state energy by about \(17.4\times\).

\begin{table*}[t]
\centering
\caption{MNIST Monte Carlo results under fixed-point wrapping arithmetic. Results are reported as mean \(\pm\) standard deviation over three runs.}
\label{tab:mnist_mc_summary}
\resizebox{\textwidth}{!}{%
\begin{tabular}{lcccccc}
\toprule
Model
& Bits
& Accuracy (\%)
& Test Loss
& Max. Energy
& Projection Rate (\%)
& Activation Overflow (\%) \\
\midrule
FP32
& --
& \(89.15 \pm 0.64\)
& 0.3491
& 6.2634
& 0.00
& 0.000 \\

FP32 + Monotone
& --
& \(86.95 \pm 0.85\)
& 0.4363
& 0.3595
& 99.98
& 0.000 \\

\midrule
PTQ Wrap
& 4
& \(11.35 \pm 1.55\)
& 2.2798
& 0.1417
& 0.00
& 0.000 \\

PTQ Wrap
& 6
& \(64.65 \pm 5.03\)
& 1.0080
& 3.8393
& 0.00
& 0.832 \\

PTQ Wrap
& 8
& \(74.77 \pm 4.09\)
& 0.7783
& 3.7592
& 0.00
& 0.796 \\

PTQ Wrap
& 12
& \(75.37 \pm 3.49\)
& 0.7656
& 3.7467
& 0.00
& 0.793 \\

\midrule
PTQ Wrap + Monotone
& 4
& \(12.50 \pm 1.78\)
& 2.2659
& 0.1233
& 99.78
& 0.000 \\

PTQ Wrap + Monotone
& 6
& \(80.83 \pm 0.84\)
& 0.6709
& 0.3800
& 99.98
& 0.041 \\

PTQ Wrap + Monotone
& 8
& \(84.10 \pm 2.18\)
& 0.5420
& 0.3602
& 99.95
& 0.039 \\

PTQ Wrap + Monotone
& 12
& \(84.02 \pm 1.95\)
& 0.5408
& 0.3587
& 99.96
& 0.040 \\

\midrule
QAT Wrap
& 4
& \(10.25 \pm 1.54\)
& 2.3037
& 0.0023
& 0.00
& 0.000 \\

QAT Wrap
& 6
& \(11.60 \pm 0.79\)
& 2.4504
& 5.5541
& 0.00
& 11.719 \\

QAT Wrap
& 8
& \(10.88 \pm 1.26\)
& 2.4365
& 5.4734
& 0.00
& 13.922 \\

QAT Wrap
& 12
& \(9.87 \pm 0.39\)
& 2.5169
& 5.4968
& 0.00
& 13.638 \\

\midrule
QAT Wrap + Monotone
& 4
& \(10.25 \pm 1.54\)
& 2.3037
& 0.0023
& 0.05
& 0.000 \\

QAT Wrap + Monotone
& 6
& \(82.40 \pm 4.32\)
& 0.5482
& 0.4585
& 100.00
& 0.012 \\

QAT Wrap + Monotone
& 8
& \(85.80 \pm 2.24\)
& 0.4694
& 0.3907
& 99.97
& 0.008 \\

QAT Wrap + Monotone
& 12
& \(86.55 \pm 0.65\)
& 0.4426
& 0.3714
& 99.99
& 0.009 \\
\bottomrule
\end{tabular}%
}
\end{table*}

The post-training quantisation results show that wrapping arithmetic degrades the unconstrained model, especially at low precision. At \(4\) bits, PTQ collapses to \(11.35\%\pm1.55\%\), close to random guessing. Increasing the precision to \(6\) bits improves accuracy to \(64.65\%\pm5.03\%\), while \(8\)--\(16\) bits reach approximately \(75\%\). However, even at high precision, unconstrained PTQ remains well below the FP32 baseline and exhibits nonzero activation overflow of approximately \(0.79\%\).

Adding monotone Lyapunov projection significantly improves PTQ robustness. At \(6\) bits, PTQ with monotone projection increases accuracy from \(64.65\%\) to \(80.83\%\), while reducing the maximum energy from \(3.8393\) to \(0.3800\). At \(8\) and \(12\) bits, the projected PTQ models reach approximately \(84\%\) accuracy and suppress activation overflow to about \(0.04\%\). This indicates that energy-bounded state evolution reduces the probability of entering overflow-prone regions of the fixed-point dynamic range.

The most important behaviour appears in the QAT experiments. Unconstrained wrapped QAT fails across \(6\)--\(16\) bits, remaining close to chance-level accuracy despite increasing precision. For example, QAT Wrap at \(8\) bits achieves only \(10.88\%\pm1.26\%\) accuracy and exhibits \(13.922\%\) activation overflow. Similar behaviour occurs at \(10\), \(12\), and \(16\) bits. This shows that the failure is not simply insufficient bit-width; rather, the learned dynamics enter regions where fixed-point wrapping corrupts the sign and magnitude of hidden activations.

In contrast, QAT with monotone Lyapunov projection remains stable and trainable. At \(6\) bits, QAT Wrap + Monotone achieves \(82.40\%\pm4.32\%\) accuracy while reducing activation overflow to \(0.012\%\). At \(8\) bits, it reaches \(85.80\%\pm2.24\%\), and at \(12\) bits it achieves the best projected QAT result of \(86.55\%\pm0.65\%\), with only \(0.009\%\) activation overflow. These results demonstrate that monotone Lyapunov projection acts as a hardware-aware stabilisation mechanism: it prevents the hidden state from entering overflow-prone regions and thereby avoids catastrophic two's-complement wraparound.

The \(4\)-bit case behaves differently. Both QAT Wrap and QAT Wrap + Monotone remain near chance accuracy, but their activation overflow is essentially zero and the maximum energy is only \(0.0023\). This indicates representation collapse rather than overflow instability. In other words, \(4\)-bit precision is too coarse for the chosen fixed-point format and model configuration; the hidden states are quantised into a near-degenerate representation before overflow becomes the dominant failure mode.

\begin{figure}[t]
\centering
\includegraphics[width=0.88\columnwidth]{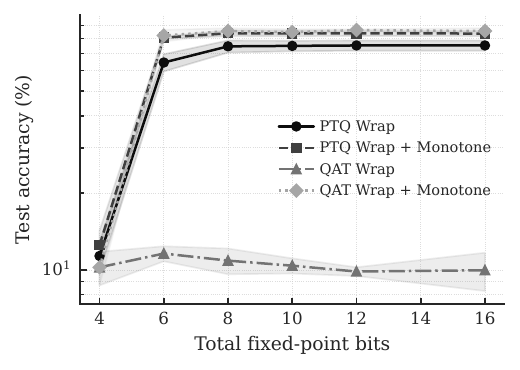}
\caption{MNIST test accuracy versus fixed-point bit-width under hardware-style wrapping. Monotone Lyapunov projection substantially improves both PTQ and QAT robustness, particularly in the \(6\)--\(12\)-bit range.}
\label{fig:mnist_quant_sweep_accuracy}
\end{figure}

\begin{figure}[t]
\centering
\includegraphics[width=0.88\columnwidth]{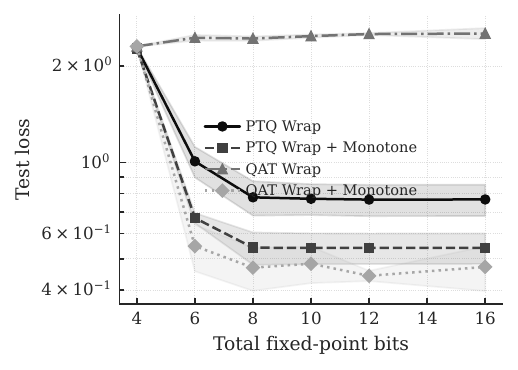}
\caption{Test loss versus fixed-point bit-width. Unconstrained wrapped QAT remains close to chance-level loss, whereas monotone projection restores stable optimisation.}
\label{fig:mnist_quant_sweep_loss}
\end{figure}

\begin{figure}[t]
\centering
\includegraphics[width=0.88\columnwidth]{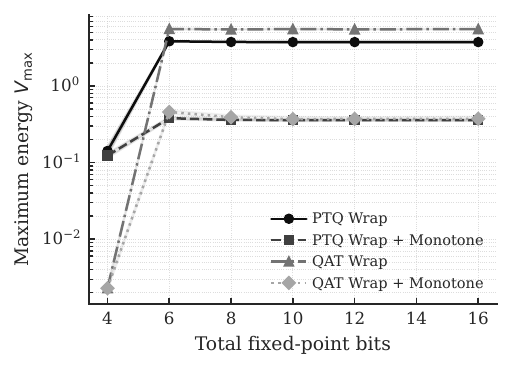}
\caption{Maximum layerwise Lyapunov energy versus bit-width. The proposed monotone projection keeps the hidden-state energy bounded across the quantisation sweep.}
\label{fig:mnist_quant_sweep_max_energy}
\end{figure}

\begin{figure}[t]
\centering
\includegraphics[width=0.88\columnwidth]{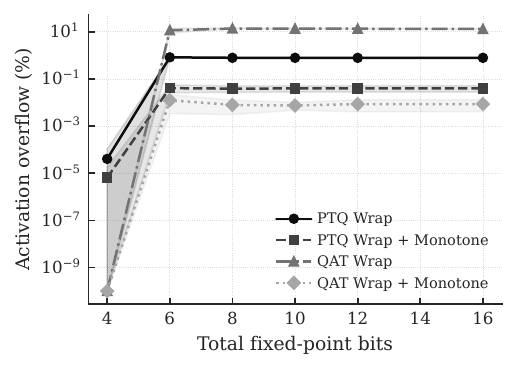}
\caption{Activation overflow rate versus fixed-point bit-width. Unconstrained wrapped QAT exhibits severe overflow, while monotone Lyapunov projection suppresses overflow to near-zero levels.}
\label{fig:mnist_quant_sweep_activation_overflow}
\end{figure}

\begin{figure}[t]
\centering
\includegraphics[width=0.88\columnwidth]{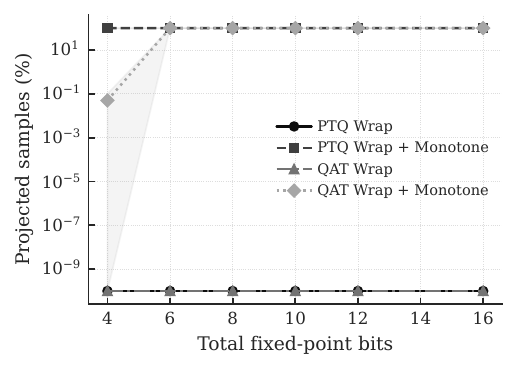}
\caption{Projection rate across the quantisation sweep. The high projection rate indicates that the Lyapunov constraint is actively enforcing a safe hidden-state set throughout the network.}
\label{fig:mnist_quant_sweep_projection_rate}
\end{figure}

\begin{figure}[t]
\centering
\includegraphics[width=0.88\columnwidth]{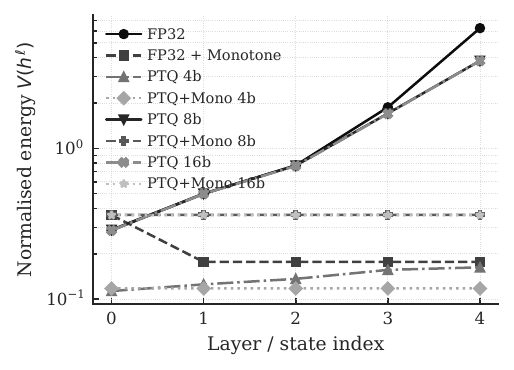}
\caption{Representative layerwise Lyapunov energy curves. The unconstrained models permit energy growth across depth, whereas monotone projection enforces bounded depth-wise dynamics.}
\label{fig:representative_layerwise_energy}
\end{figure}

\begin{figure}[t]
\centering
\includegraphics[width=0.88\columnwidth]{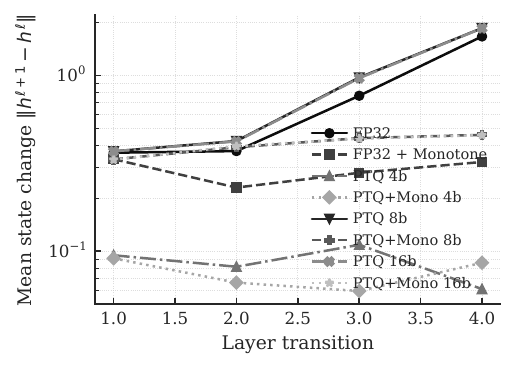}
\caption{Representative layerwise state-change curves. Projection reduces the magnitude of inter-layer perturbations, indicating more controlled hidden-state evolution under fixed-point arithmetic.}
\label{fig:representative_layerwise_delta}
\end{figure}

Overall, the results support three conclusions. First, standard FP32 training provides the best unconstrained accuracy but allows substantial hidden-state energy growth. Second, hardware-style wrapping is a severe failure mode for low-precision transformer dynamics; unconstrained wrapped QAT collapses to chance-level accuracy even at high bit-widths because overflow corrupts the hidden states. Third, monotone Lyapunov projection restores stable fixed-point learning by bounding the hidden-state energy and suppressing activation overflow. The best projected QAT result is obtained at \(12\) bits, where the model achieves \(86.55\%\pm0.65\%\) accuracy with only \(0.009\%\) activation overflow.

\section{Conclusion}
\label{sec:conclusion}

This paper investigated the stability of low-precision neural networks under realistic fixed-point arithmetic with two's-complement overflow wrapping. Unlike idealised clipping or saturation models, wrapping arithmetic can map large positive values to negative values, and vice versa, thereby corrupting the hidden-state dynamics of a neural network. To address this problem, we introduced a Lyapunov-based monitoring and projection mechanism that constrains the normalised hidden-state energy across network depth. In particular, the monotone projection enforces
\[
V(h^{\ell+1}) \leq \min\{V(h^\ell),V_{\max}\},
\]
thereby ensuring bounded and non-increasing layerwise energy.

The MNIST experiments demonstrate that hidden-state energy growth is a major source of failure in low-precision wrapped arithmetic. The unconstrained full-precision model achieved \(89.15\%\pm0.64\%\) accuracy, but exhibited a maximum normalised hidden-state energy of \(6.2634\). Applying monotone Lyapunov projection reduced the maximum energy to \(0.3595\), with a modest accuracy reduction to \(86.95\%\pm0.85\%\). This confirms that the proposed constraint provides a meaningful stability--accuracy trade-off.

The most significant gains were observed under quantisation-aware training with wrapping arithmetic. Without Lyapunov projection, wrapped QAT collapsed to near-chance performance across \(6\)--\(16\) bits, with activation overflow rates between approximately \(11.7\%\) and \(13.9\%\). This shows that increasing bit-width alone does not necessarily prevent failure when the internal dynamics remain unconstrained. By contrast, QAT with monotone Lyapunov projection suppressed activation overflow to almost zero and restored stable learning. The best result was obtained at \(12\) bits, where the projected model achieved \(86.55\%\pm0.65\%\) accuracy with only \(0.009\%\) activation overflow.

These findings suggest that Lyapunov-based projection can serve as a hardware-aware safety mechanism for low-precision neural networks. Rather than treating quantisation error only as a numerical approximation problem, the proposed approach views low-precision inference and training as a dynamical stability problem. By constraining the hidden-state trajectory, the method prevents the network from entering overflow-prone regions of the fixed-point state space.

Future work will extend the analysis to larger datasets, deeper transformer architectures, and hardware-in-the-loop FPGA or ASIC implementations. Further work will also investigate adaptive thresholds, layer-dependent safe sets, and joint optimisation of accuracy, energy consumption, overflow probability, and Lyapunov stability margins.


\printbibliography
\end{document}